%% file: main.tex
\documentclass[10pt,twocolumn,letterpaper]{article}

\usepackage{iccv}              % To produce the CAMERA-READY version
\usepackage[accsupp]{axessibility}

\definecolor{iccvblue}{rgb}{0.21,0.49,0.74}
\usepackage[pagebackref,breaklinks,colorlinks,allcolors=iccvblue]{hyperref}

\usepackage{longtable, booktabs}
\usepackage{multirow}
\usepackage[acronym]{glossaries}
\glsdisablehyper
\newacronym{svcs}{SVCS}{Surround-view Camera System}
\newacronym{vcam}{VCAM}{Virtual Camera}
\newacronym{uls}{ULS}{Ultrasonic Level Sensor}
\newacronym{uwb}{UWB}{Ultra-Wideband}
\newacronym{bev}{BEV}{Bird Eye View}
\newacronym{mjpe}{MPJPE}{Mean Per Joint Position Error}
\newacronym{fov}{FoV}{Field of View}
\newacronym{adas}{ADAS}{Advanced Driver-Assistance System}
\newacronym{vnf}{VNF}{Valeo Near-Field}
\newacronym{ade}{ADE}{Average Displacement-Error}

\def\paperID{5} % *** Enter the Paper ID here
\def\confName{ICCV}
\def\confYear{2025}

\title{Valeo Near–Field: a novel dataset for pedestrian intent detection}

\author{Antonyo Musabini, Rachid Benmokhtar, Xavier Perrotton\\
Valeo, BRAIN Division\\
6 rue Daniel Costantini 94000\\
Créteil - France\\
{\tt\small antonyo.musabini@valeo.com}
\and
Jagdish Bhanushali \\
Valeo, BRAIN Division\\
2100 South El Camino Real, suite D100\\
San Mateo - CA 94403, United States\\
{\tt\small jagdish.bhanushali@valeo.com}
\and
Victor Galizzi, Bertrand Luvison\\
Universite Paris-Saclay, CEA, List,\\
 F-91120, Palaiseau, France,\\
{\tt\small bertrand.luvison@cea.fr}
}

\begin{document}
\maketitle
\input{sec/0_abstract}    
\input{sec/1_intro}
\input{sec/2_related_work}
\input{sec/3_dataset}
\input{sec/4_tasks}
\input{sec/5_conclusion}
\input{sec/6_Acknowledgments}
{
    \small
    \bibliographystyle{ieeenat_fullname}
    \bibliography{main}
}

\end{document}

%% file: sec/0_abstract.tex
\begin{abstract}
This paper presents a novel dataset aimed at detecting pedestrians' intentions as they approach an ego-vehicle. The dataset comprises synchronized multi-modal data, including fisheye camera feeds, lidar laser scans, ultrasonic sensor readings, and motion capture-based 3D body poses, collected across diverse real-world scenarios. Key contributions include detailed annotations of 3D body joint positions synchronized with fisheye camera images, as well as accurate 3D pedestrian positions extracted from lidar data, facilitating robust benchmarking for perception algorithms. We release a portion of the dataset along with a comprehensive benchmark suite, featuring evaluation metrics for accuracy, efficiency, and scalability on embedded systems. By addressing real-world challenges such as sensor occlusions, dynamic environments, and hardware constraints, this dataset offers a unique resource for developing and evaluating state-of-the-art algorithms in pedestrian detection, 3D pose estimation and 4D trajectory and intention prediction. Additionally, we provide baseline performance metrics using custom neural network architectures and suggest future research directions to encourage the adoption and enhancement of the dataset. This work aims to serve as a foundation for researchers seeking to advance the capabilities of intelligent vehicles in near-field scenarios. % The public test-set is available on \url{https://github.com/antonyomusabini/Valeo_NearField}
\end{abstract}

%% file: sec/1_intro.tex
\section{Introduction}
\label{sec:intro}

As intelligent vehicles continue to advance, ensuring safe and effective interaction with pedestrians in various environments remains a key challenge. A critical aspect of this interaction is the ability to predict and understand a pedestrian's intent as they approach a parked, static ego-vehicle. This capability is essential for anticipating their intentions through novel \gls{adas}. For instance, a vehicle could adjust its seats and mirrors to match the morphology of an approaching pedestrian or automatically open doors or the trunk if the pedestrian is authorized. These \gls{adas} must ensure that the vehicle avoids collisions with automatically opened doors and does not introduce any safety risks.

\begin{figure}[htbp]
    \centerline{\includegraphics[trim=0 0 500 0,clip, width=\linewidth]{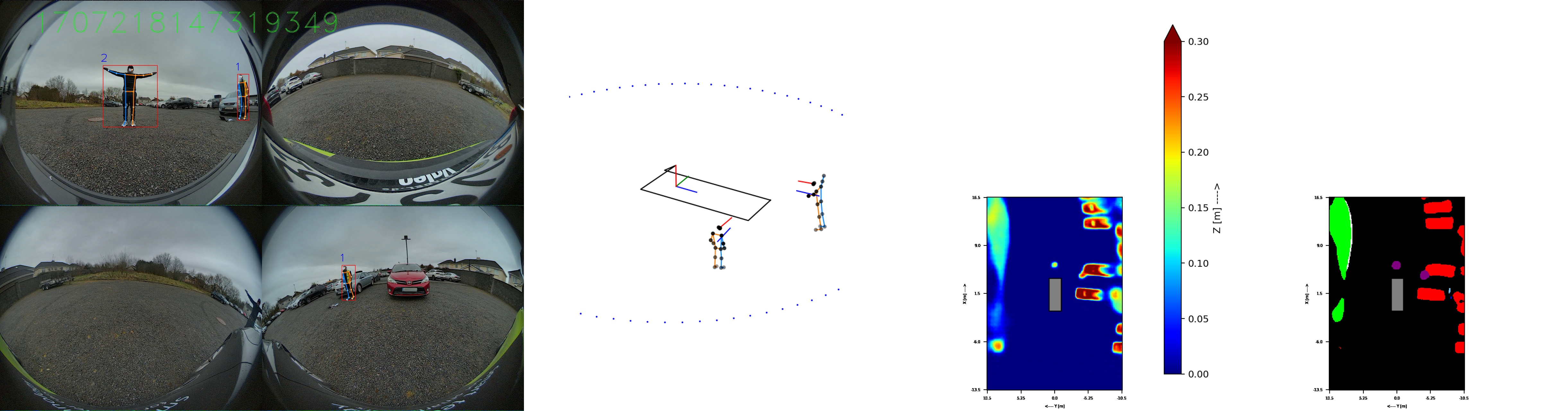}}
    \caption{\textbf{Valeo Near Field Dataset.} Inference visualization on an exemplary scene. The four fisheye streams are shown on the left side, displaying the detected 2D body joints, 2D bounding boxes, and camera-consistent IDs for each pedestrian. On the right side, a rectangular representation of the vehicle is depicted along with the 3D pose estimations of both pedestrians, including head orientation vectors and upper-body orientation vectors. Additionally, a dotted circle highlights the 10m range around the vehicle.}
    \label{fig::cover_image}
\end{figure}

Current datasets for pedestrian detection and tracking are often limited in scope, typically focusing on basic pedestrian detection or static poses without addressing the challenges of real-world dynamic interactions between pedestrians and vehicles. Many existing datasets emphasize pedestrians at crossings and do not specifically address scenarios involving a parked ego-vehicle and related use cases. Additionally, many datasets lack the multimodal sensor integration necessary to capture the complex nature of pedestrian behavior in urban environments. The absence of high-quality, synchronized multimodal data with relevant intention scenarios hinders the development and benchmarking of robust pedestrian detection and intent prediction algorithms in the near field of vehicles.

In this paper, we present a novel dataset specifically designed to address these challenges. Our dataset includes 300 sequences with synchronized multimodal data from fisheye cameras, LiDAR laser scanners, ultrasonic sensors, and motion capture systems, all collected in diverse real-world environments. Each sequence lasts, on average, 1 minute and is captured at 30 frames per second, resulting in a total of 540,000 frames. Executed by 13 participants, this rich dataset enables the accurate estimation of pedestrian positions, 3D body joint locations, and intent prediction. By providing publicly 51 sequences from a total of 12 participants, 42 outdoor and 9 indoor scenes, as \textit{\gls{vnf} public-test set}, we aim to foster the development of more robust algorithms that can accurately predict pedestrian intent, even in the presence of occlusions and dynamic environmental changes. Additionally, we offer a comprehensive benchmark suite, allowing researchers to evaluate and compare the performance of their algorithms.
 
This work serves as both a resource and a benchmark for advancing pedestrian detection and intent prediction, ultimately contributing to the safe deployment of autonomous vehicles in complex, real-world scenarios.

%% file: sec/2_related_work.tex
\section{Related Work}
\label{sec:RelatedWork}

\subsection{Applications}

Human body pose-related applications can be categorized into four main groups based on their input and output data formats:

\begin{itemize}
    \item \textbf{2D Body Joints}: This category includes models that take a single image as input and output detected and tracked 2D body joints of pedestrians in the scene: in real-time \cite{jiang2023rtmposerealtimemultipersonpose}, for whole-body pose estimation, including not just the major joints (like elbows, wrists, knees, and ankles) but also finer details such as facial landmarks, hand keypoints, and foot keypoints \cite{xu2022vitpose}, and as a single-stage detection and tracking capability \cite{kreiss2021openpifpaf}. It also encompasses works that infer additional attributes for safety-critical applications in autonomous vehicles based on 2D poses, such as a pedestrian’s walking intention or gender \cite{9527172}, as these tasks remain within the image domain.

    \item \textbf{3D Pose Estimation}: This category includes methods that take 2D body joints as input and predict an uplifted 3D pose using geometry-driven attention mechanism to address uncalibrated multi-view cameras \cite{10581862}, transformers \cite{mehraban2024motionagformer, motionbert2022}, diffusion models \cite{rommel2023diffhpe} or temporal convolutions \cite{pavllo20193dhumanposeestimation}. The estimated poses are ego-centric, meaning they do not account for the actual distance between the pedestrian and the camera.

    \item \textbf{3D Localization}: This category consists of approaches that estimate the actual distance between pedestrians and the camera, placing them correctly within a 3D scene. These methods leverage various features, such as 3D projections of foot positions across multiple camera views \cite{lima2021generalizable, hou2020multiview} or the detected 2D height of pedestrians \cite{bertoni2019monoloco} for monocular 3D pedestrian localization and uncertainty estimation.

    \item \textbf{4D Trajectory Prediction}: This category focuses on predicting socially plausible pedestrian trajectories in crowded environments. Some approaches rely solely on 3D localization \cite{STGAT}, while others incorporate full 3D pose information for trajectory forecasting \cite{saadatnejad2024socialtransmotion, gao2024multi}.

    \item \textbf{4D Pedestrian Intent Detection}: This category includes works that detect pedestrian intentions, either from video sequences or 3D pose estimation. For instance, VRUNet \cite{VRUNet} predicts pedestrian intentions while they walk or cross urban roads, with particular attention to pedestrians at crosswalks.

\end{itemize}

\subsection{Datasets}

% MOT17 Dataset, KITTI Dataset, EuroCity Persons Datase, CityPersons Dataset

The common datasets used for image-based methods are the following:

COCO Keypoint Dataset \cite{dai2020dark} - part of the larger COCO dataset, which also includes object detection and segmentation tasks - contains annotations for 17 keypoints on the body from the image domain. CrowdPose \cite{li2019crowdpose}, an extension of COCO, focuses on crowded scenes where people are in close proximity, resulting occlusions and overlapping bodies. PoseTrack \cite{iqbal2017posetrack} is a dataset for multi-person pose estimation and tracking in videos adressing to temporal consistency in pose tracking and the challenges of handling occlusions over time. WILDTRACK \cite{chavdarova2017wildtrack} provides multi-camera annotations for multi-person pose estimation and tracking in unconstrained environments both indoor and outdoor scenes. MPII Human Pose Dataset \cite{andriluka14cvpr} provides annotations of up to 16 body joints for human activities and poses from YouTube videos, which is the source of it's real-world nature.

The common datasets used for the presented 3D Pose Estimation applications are the following:
Human 3.6M \cite{ionescu2013human3}, a very large-scale dataset, provides annotations from 3D keypoints captured using a motion capture system. Pedestrians and Cyclists in Road Traffic Dataset \cite{kress_2021_4898838} provides 3D body joint positions of vulnerable road users and semantic maps reflecting the surrounding scene. This combination is particlarly intersting for autonomous driving and pedestrian safety applications. However, the corresponding video stream are not publicly shared.

Trajectory Prediction applications use the following datasets:
SGAN \cite{sgan} is a GAN-based generated human position dataset, without real-world data. JTA (Joint Track Auto) \cite{fabbri2018learning} is a dataset for people tracking in urban scenarios by exploiting a photorealistic videogame. JRDB \cite{jrdb_dataset} is a real-world dataset that provides a diverse set of pedestrian trajectories and 2D bounding boxes. ETH-UCY Dataset \cite{ETH_BIWI_Walking_Pedestrians}  used for evaluating the social interactions and group dynamics of pedestrians, but does not contain visual cues.

% real MPT dataset.
% \cite{social_attention}
None of the previously described datasets provide the necessary training data for models designed to address the near-field detection problem around vehicles. First, all datasets designed for the \textbf{2D body joints} task exclusively use pinhole cameras, which do not account for fisheye geometric distortions affecting pedestrians located at the edges of images. Second, these datasets primarily treat the \textbf{2D body joints} detection problem as a monocular task, whereas vehicle-mounted cameras have significant overlapping fields of view, enabling the possibility of cross-view processing. Third, the vast majority of datasets used for \textbf{3D pose estimation} algorithms provide 3D poses in highly controlled laboratory environments. In these cases, the pedestrian is often fully visible and positioned at the center of the cameras’ field of view. Moreover, the operating range of these datasets is typically only a few meters, which is significantly shorter than the 15–20 meters required for vehicle-based applications. Fourth, datasets tailored for \textbf{trajectory prediction} primarily focus either on pedestrian crossings or on modeling "socially" plausible trajectories in crowded scenes. They entirely overlook the interaction of an approaching pedestrian with a stationary ego-vehicle.

To the best of our knowledge, the \textbf{Valeo Near-Field} dataset is the first publicly available dataset that provides four fisheye camera streams from a vehicle along with the captured 3D body pose from a motion suit of an approaching pedestrian with a specific intent toward the ego-vehicle. This dataset can be used for any of the presented pedestrian detection tasks individually or as an end-to-end solution for all of them simultaneously.

%% file: sec/3_dataset.tex
\section{The Valeo Near-Field Dataset}
\label{sec:VNFD}

\subsection{Data Collection Setup}

\begin{figure}[!htb]
    \centering
    \subfloat[\centering BMW G11 S7 Vehicle.]{\includegraphics[trim=0 0 200 50, clip,height=3.5cm]{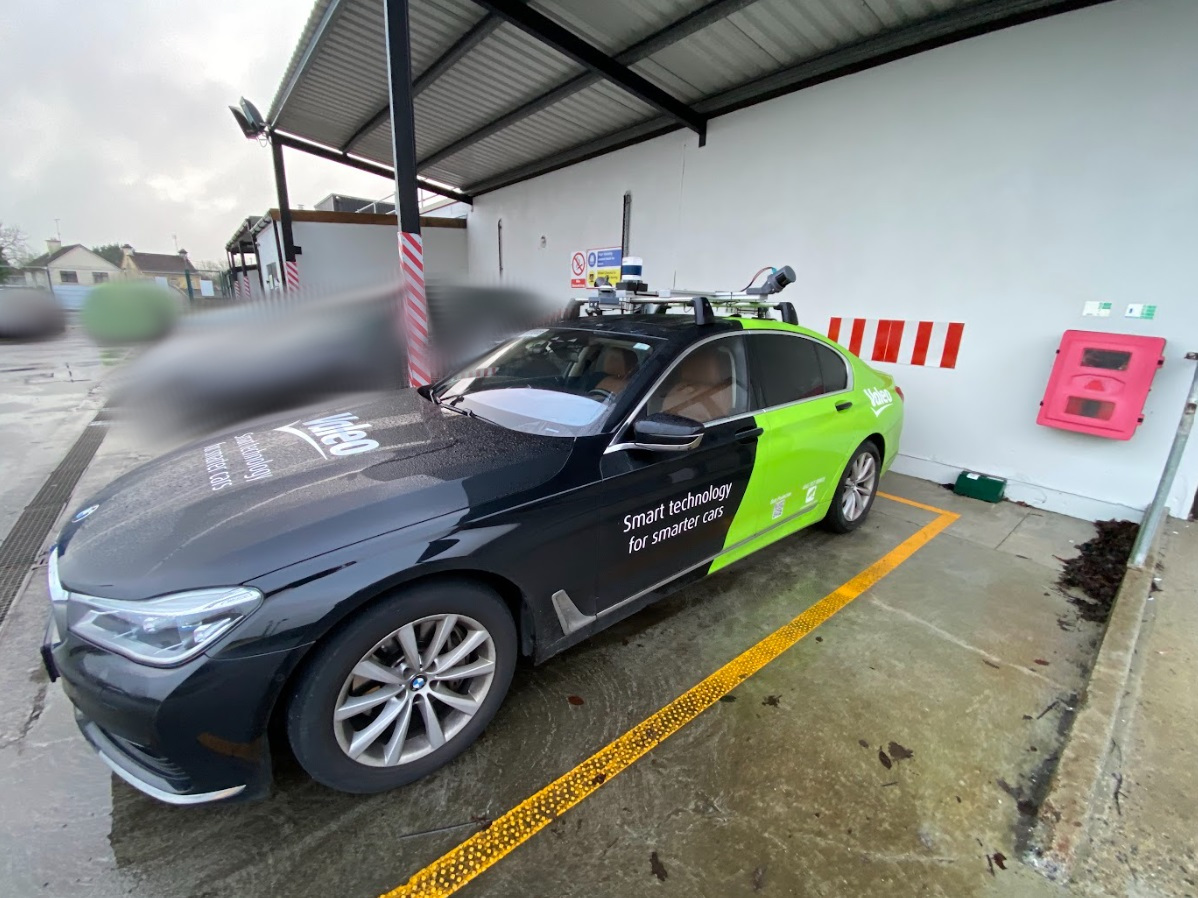}}\label{fig:S7_v1}
    \qquad
    \subfloat[\centering Positions of four laser scanners.]{\includegraphics[trim=0 0 10 50, clip,height=3.5cm]{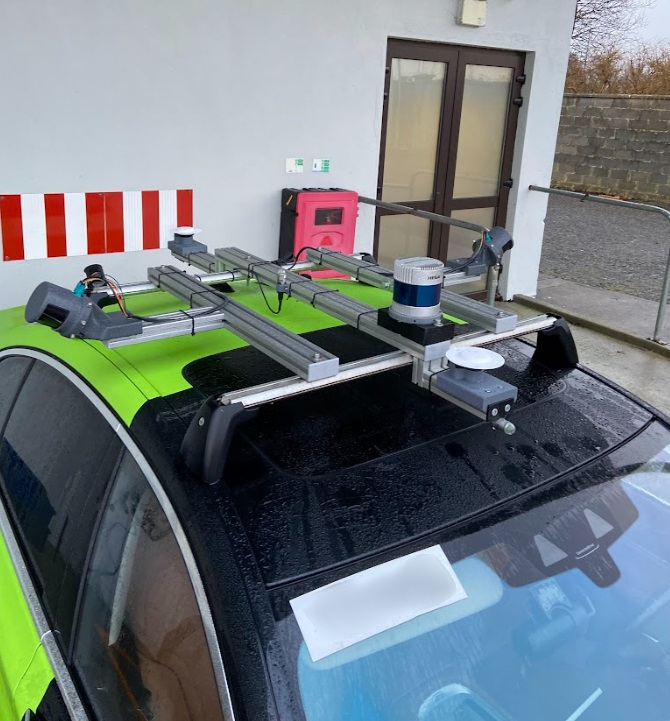}\label{fig:S7_v2}}
    \qquad
    %\vspace{-10pt}
    \caption{\textbf{Vehicle and LiDAR configuration.}
    }
    \label{fig:Vehicle}
\end{figure}

The data collection process for this dataset was designed to capture the interactions between pedestrians and a stationary ego-vehicle, simulating real-world urban scenarios. The primary objective was to gather a wide variety of multimodal sensor data to support the development and evaluation of algorithms for pedestrian detection, intent prediction, and 3D pose estimation. To achieve this, a comprehensive sensor suite was employed, including a \gls{svcs} with fisheye cameras, LiDAR scanners, ultrasonic sensors, and a motion capture system. These sensors were selected to provide complementary information about the environment and the pedestrians' behavior around the vehicle. A BMW G11 S7 was used for this data collection (see \cref{fig:S7_v1}).

\subsubsection{Sensor Suite} \label{subsubsection::SensorSuite}

\begin{figure}[!htb]
    \centering
    \subfloat[\centering Top View of the point clouds.]{{\includegraphics[,height=.3\linewidth]{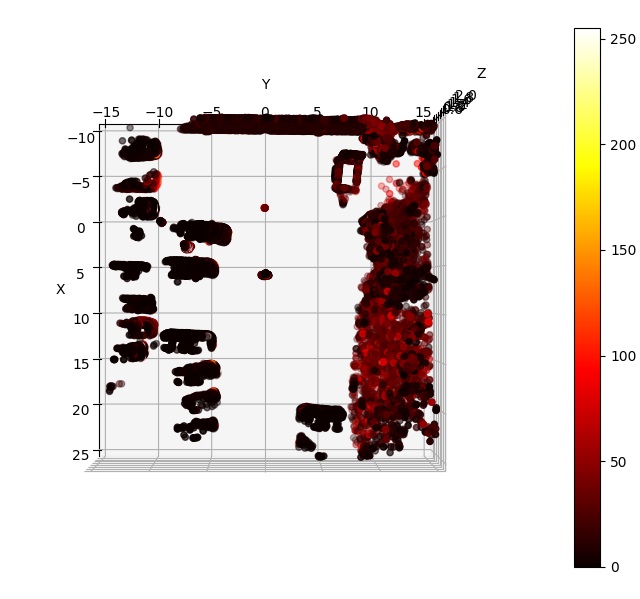}}\label{fig:annotation_pcd}}
    \qquad
    \subfloat[\centering SVCS Images.]{{\includegraphics[height=.3\linewidth]{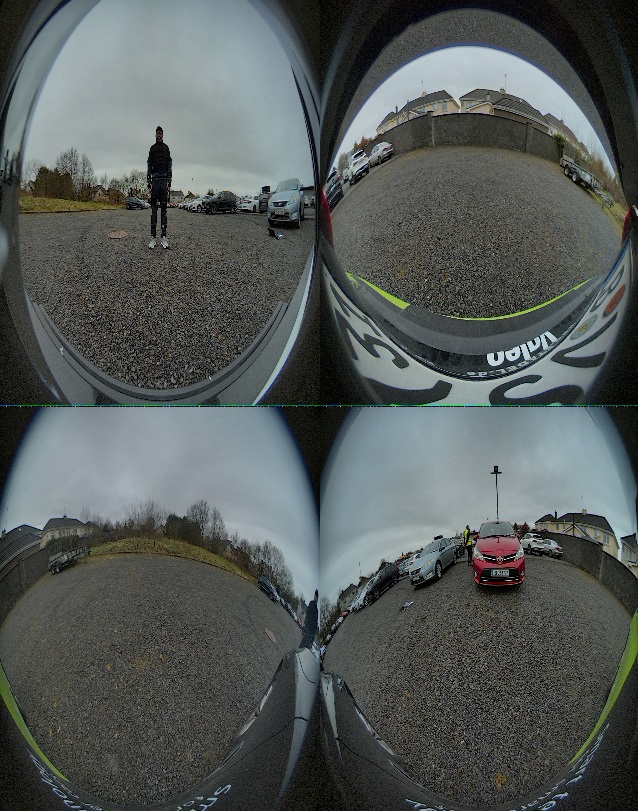}}\label{fig:annotation_image}}
    \qquad
    \subfloat[\centering 23 Pedestrian Joints (in black)~\cite{karatsidis2016estimation}.]{{\includegraphics[height=.3\linewidth]{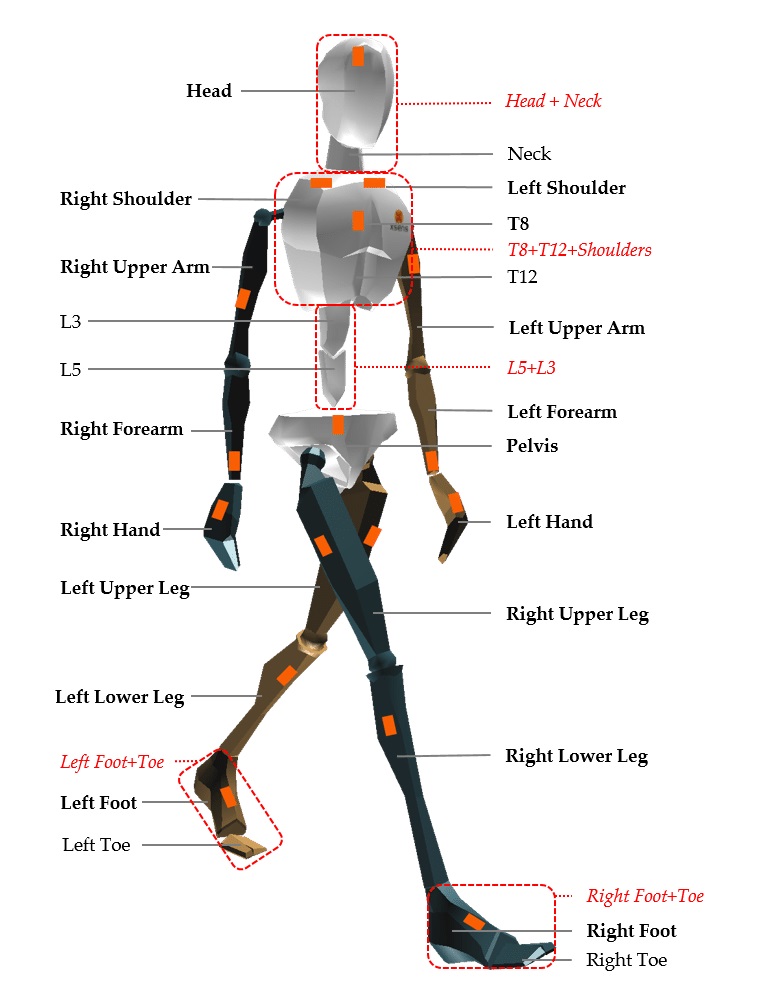}}\label{fig:mvn_full}}
    \qquad
    \caption{\textbf{Annotation Strategy.} Various elements used to create 3D skeleton annotations are visible.}
    \label{fig:annotation_trio}
\end{figure}

\begin{itemize}
    \item \gls{svcs} \textbf{fisheye cameras} are typically installed at the front, rear, left, and right of vehicles and are commonly known as parking cameras. Their strategic placement around the vehicle allows for capturing a 360-degree \gls{fov} without leaving any blind spots. The fisheye camera feeds are synchronized with the other sensors to ensure accurate multimodal fusion and data alignment. A frame from the \gls{svcs} is shown on the right side of \cref{fig::cover_image}.

    \item A set of four \textbf{LiDAR} sensors was used to capture precise distance measurements of the surrounding environment, even for the near-field (see \cref{fig:S7_v2} for the sensor position). This modality is particularly effective for accurately annotating pedestrians' positions, as one single top mounted sensor would suffer from occlusions created from ego-vehicle body for the very close range.

    \item \textbf{Ultrasonic sensors} were deployed to detect proximity and measure the distance between the vehicle and nearby objects, including pedestrians. These sensors are especially useful for detecting pedestrians at close range, such as when they approach the vehicle's door or trunk.

    \item An MVN Awinda \cite{MVN_Awinda} \textbf{motion capture suit} was used to capture the 3D body poses of pedestrians as they moved around the vehicle. The motion capture data provides ground truth for the 3D body joint positions, offering a detailed view of pedestrian behavior. The captured 23 joints are shown in \cref{fig:mvn_full}.

\end{itemize}

\subsubsection{Synchronization of Multimodal Data} \label{subsubsection::Synchronization}

\begin{figure}[!htb]
    \centering
    \subfloat[\centering Stand Still Pose]{\includegraphics[trim=200 30 100 30, clip, height=3.2cm]{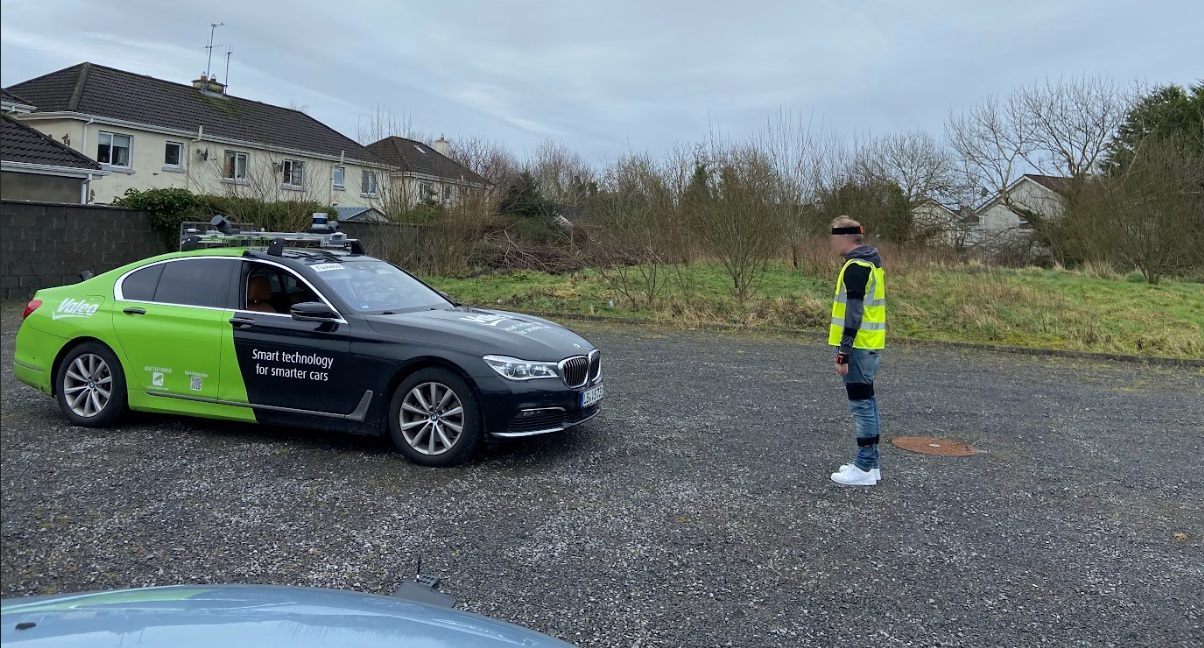}\label{fig:calib_0}}
    \qquad
    \subfloat[\centering Synchronization Trigger]{\includegraphics[trim=220 55 100 15, clip,height=3.2cm]{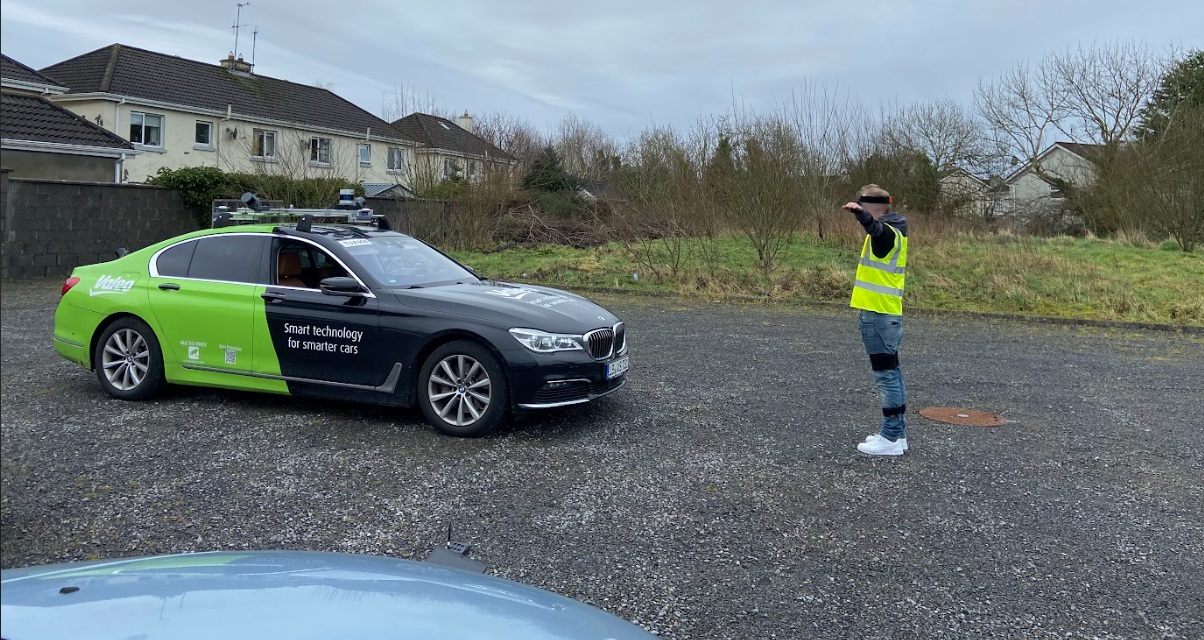}\label{fig:calib_1}}
    \qquad
    %\vspace{-10pt}
    \caption{\textbf{Vehicle and sensor configuration.}
    }
    \label{fig:SynchMouvement}
\end{figure}

One of the key challenges in collecting this dataset was ensuring the synchronization of perfectly aligned vehicle sensors (fisheye cameras, LiDARs and ultrasonic sensors) with the motion suit. This allowed for accurate cross-referencing of the data across different sensor modalities, facilitating the development of multimodal perception algorithms.

Due to the lack of timestamps that could synchronize the motion suit data with the remaining sensors, we developed a tailor-made synchronization technique. According to our protocol, the pedestrian first adopts a stationary pose at any position next to the vehicle (see \cref{fig:calib_0}). Sequentially, the left arm is raised and lowered back to the initial position (see \cref{fig:calib_1}), followed by the right arm being raised and lowered, and finally, both arms are raised and lowered simultaneously. These specific movements are automatically matched across the image space (captured by fisheye cameras) and the motion suit data space to establish time synchronization between the sensor sets. The 3D position at which this movement is performed (which may vary for each scene) is designated as the motion suit root position.

The motion suit used in our setup operates with an open-loop 3D position registration. This implies that errors in the pedestrian’s 3D position may accumulate over time between the start of data acquisition at the root position and the end of the sequence, regardless of the accuracy of the 3D pose itself. To mitigate this drift, we manually annotated the pedestrian's 3D position and orientation (from a top-view perspective) using LiDAR scans at a frequency of 3 fps (see \cref{fig:annotation_pcd}). The time-synchronized motion suit data was then aligned with these manual annotations, effectively minimizing the impact of accumulated positional errors.

Finally the motion suit data is translated to vehicle coordinate system, which the root position is the center of the rear axle of the vehicle.

\begin{figure}[!htb]
    \centering
    \subfloat{{\includegraphics[height=1.8cm]{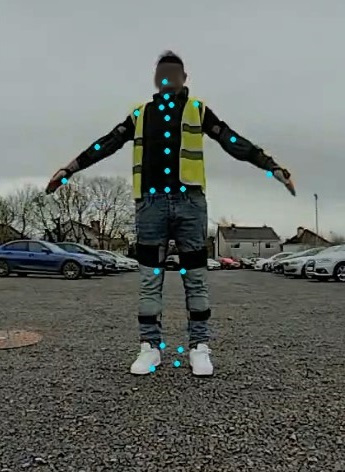}}\label{fig:x_sens_on_image:0}}
    \qquad
    \subfloat{{\includegraphics[trim=10 0 10 0, clip, height=1.8cm]{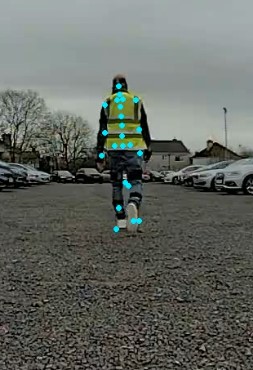}}\label{fig:x_sens_on_image:1}}
    \qquad
    \subfloat{{\includegraphics[height=1.8cm]{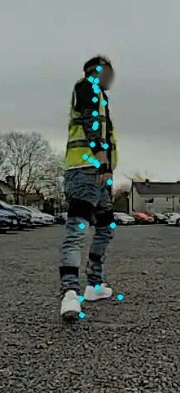}}\label{fig:x_sens_on_image:2}}
    \qquad
    \subfloat{{\includegraphics[height=1.8cm]{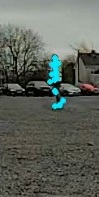}}\label{fig:x_sens_on_image:3}}
    \qquad
    \subfloat{{\includegraphics[height=1.8cm]{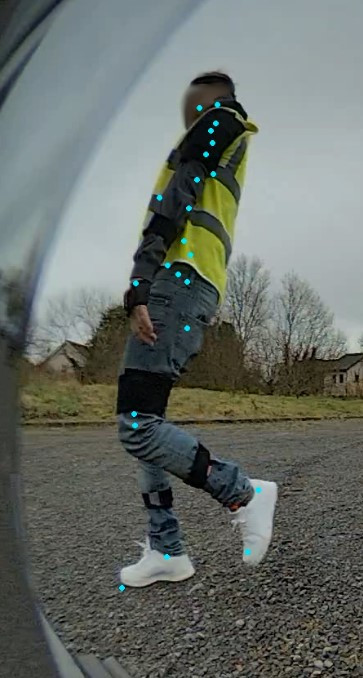}}\label{fig:x_sens_on_image:4}}
    \qquad
    \caption{\textbf{Annotation Visualization.} Projection of 3D skeleton data on a 2D fisheye image.}
    \label{fig:x_sens_on_image}
\end{figure}

Figure \ref{fig:x_sens_on_image} illustrates a couple of examples of the 3D skeleton data obtained from the motion suit. Once synchronized in time and corrected with translation and rotation as described above, the motion suit data are projected onto the fisheye images to create the illustrations in cyan circles, as a sanity check.

\subsubsection{Data Collection Process} \label{subsubsection::DataCollection}

Thirteen participants volunteered for the creation of the data set (10 males, 3 females). The mean height of the participants is 174.6 cm ($\pm$5.5cm), with the shortest at 164 cm and the tallest at 182 cm. Participants executed both intent and non-intent scenarios. Some examples other than the regular intent and non-intent use cases are as followings: placing a box in the trunk, placing a shopping basket to rear before sitting to front, talking with phone next to the vehicle, approaching to the vehicle to only talk with the driver, triggering gestures executed while walking to the vehicle. Each scenario starts with the synchronization movement of the motion suit (see \cref{subsubsection::Synchronization}). Then the pedestrian(s) move to the starting position, usually 15-20 meters from the vehicle, and finally they execute the scenario itself. In an intent scenario, the pedestrian performed actions such as opening one of the doors, sitting in the vehicle and closing the door, or opening, interacting, and closing the trunk. In a non-intent scenario, the pedestrian walked past the vehicle as a random passerby. The scenarios involved the participation of 1 to 3 pedestrians. To ensure the dataset captures realistic pedestrian behavior in different conditions, data was collected both in urban outdoor and indoor parking lots.

All participants provided written consent for data collection. To ensure the privacy of volunteers, all visible faces and license plates in the \gls{vnf} public test set have been blurred.

%% file: sec/4_tasks.tex
\section{Experiments \& Baseline Results}
\label{sec:experiments}

\subsection{Experiments}  
This section describes the state-of-the art methods used to accomplish common 3D keypoint-based applications. First, we employed YOLOX \cite{ge2021yolox} to detect pedestrians from fisheye images. To estimate 2D skeletons, the detected regions of interest within the bounding boxes were processed using our fisheye geometry-aware, fine-tuned version of ViTPose \cite{xu2022vitpose}. The detected 2D keypoints were then projected from the fisheye camera geometry onto a pinhole camera geometry to obtain the final 2D keypoints.  

Next, these 2D keypoints were uplifted to 3D pose estimation using the method described in \cite{10581862}. This two-stage approach first estimates a pedestrian's 3D pose from the 2D keypoints detected in each camera view. Then the \textit{pedestrian's 3D location is estimated} assuming a constant height. This is a naive assumption that is adjusted when pedestrian is visible from multiple cameras by associating independent 3D pose estimates corresponding to the same person and merging them by crossing line of sight. The association is done minimizing a bipartite graph cost whose cost matrix represents the distance between the independent 3D pose estimates expressed in the same world coordinates.

Once the association of individual 3D poses from each camera view is completed using the \textit{estimated} 3D locations from the previous method, a pedestrian-adapted version of the approach from \cite{doi:10.1049/icp.2024.3273} is applied to precisely \textit{predict} the actual 3D locations of pedestrians. This transformer-based method explicitly accounts for the geometry of multi-view fisheye cameras, enabling accurate positioning of pedestrian. Finally, the associated 3D pose and predicted 3D locations are matched based on the Euclidean distance between the \textit{estimated} and \textit{predicted} 3D locations.

It is worth mentioning that the described pipeline consists of multiple successive models. The results presented in this paper are obtained through direct inference on the \gls{vnf} dataset. This provides an experimental and methodological baseline for future research and development using this dataset.

\subsection{Metrics}

Due to the nature of the presented tasks, the detection performance of algorithms on the \gls{vnf} dataset is evaluated across different detection zones, each spanning 5 meters. The first zone covers the very close range (0 to 5 meters), where parts of the pedestrian may be obstructed by the ego-vehicle's body. The second zone (5 to 10 meters) ensures full pedestrian visibility while maintaining a reasonable distance from the fisheye cameras, optimizing image clarity. The remaining zones extend from 10 to 15 meters and 15 to 20 meters, allowing for performance analysis at increasing distances (task-depending).

3D pose detection algorithms are evaluated using the \gls{mjpe}, which measures the mean joint position error over the 3D body poses of pedestrians. Since \gls{mjpe} does not account for the actual height of the pedestrian, we introduce a secondary metric, \textit{Body Height Error}, to assess discrepancies in height estimation.

Pedestrian localization is assessed using the \textit{accuracy} and \gls{ade}. For each ground-truth pedestrian, the closest prediction is matched using the Hungarian algorithm, with a dynamic distance threshold set to 10\% of the pedestrian’s distance from the ego vehicle (minimum threshold of 1 m). Accuracy is defined as the ratio of correctly matched ground-truth points to the total number of ground-truth points. The \gls{ade} is computed for matched points as the distance between predicted and ground-truth pelvis positions, evaluated separately for each distance zone.

In most object detection tasks, if an object is occluded beyond a certain threshold, it is excluded from evaluation. However, in this use case, pedestrians may become partially or completely occluded. For example, a pedestrian standing too close to the vehicle may have only certain body parts visible in the camera view, or an approaching pedestrian may pass behind another vehicle, resulting in full occlusion. Nevertheless, the \gls{mjpe} and \gls{ade} should still be computed as if the pedestrian were fully visible. This can be achieved by logically completing the missing body parts through pose completion techniques and, for instance, by tracking the movement of approaching pedestrians.

\subsection{Baseline Results}

The baseline results were calculated on \textit{\gls{vnf} public-test set} (see Section \ref{sec:intro}). All evaluations focus exclusively on the main pedestrian in each scene. Global results are reported as the mean over the entire  \textit{\gls{vnf} public-test set}. For use-case-specific results (outdoor/indoor, male/female), the reported value is the average of the mean results computed per participant.

\begin{table*}[!htb]
    \centering
    \caption{\textbf{3D Pedestrian Positioning.} Average Distance Error on indoor / outdoor and male / female computed on \textit{\gls{vnf} public-test set}.}
    \label{tab:position_error_use_cases}
\begin{tabular}{c|c|c|cccc}
\hline
\multirow{2}{*}{Pred. Dist.} & \multirow{2}{*}{Accuracy} &\multicolumn{5}{c}{Average Distance Error (m)}   \\ \cline{3-7}
 & & \textbf{Mean Absolute Error} ($\pm$std)  & Outdoor & Indoor & Male & Female \\ \hline
0 to 5 m & 0.62 &  \textbf{0.32} ($\pm$0.08)  & 0.33 & 0.34 & 0.32 & 0.39  \\ \hline
5 to 10 m & 0.86 & \textbf{0.32} ($\pm$0.08)  &  0.34 & 0.37 & 0.35 & 0.33 \\ \hline
% Count (0 to 5 m): 1709, (5 to 10 m) 2169
\end{tabular}
\end{table*}

\begin{table*}[!htb]
    \centering
    \caption{\textbf{Skeleton Joint Results.} Computed \gls{mjpe} per distance zone on \textit{\gls{vnf} public-test set}.}
    \label{tab:mjpe}
    \begin{tabular}{c|c|cc|cc} \hline 
         % Prediction Distance & Total \gls{mjpe} (mm) \\ \hline 

         \multirow{2}{*}{Prediction Distance}& \multicolumn{5}{c}{\gls{mjpe} (mm)}   \\ \cmidrule(lr){2-6}
         & \textbf{Mean Total Error} ($\pm$std) & Outdoor & Indoor & Male & Female  \\ \hline 
         
         0 to 5 m & 201 $(\pm 32)$ & 211 & 205 & 204 & 221 \\ \hline 
         5 to 10 m &  180 $(\pm 44)$ & 174 & 204 & 183 & 195 \\ \hline 
         10 to 15 m & 183 $(\pm 42)$ & 177 & 213 & 183 & 223 \\ \hline 
         15 to 20 m & 199 $(\pm 35)$ & 208 & 212 & 210 & 208 \\ \hline
    \end{tabular}
\end{table*}

\begin{table*}[!htb]
    \centering
    \caption{\textbf{Body Height Results.} Errors on the predicted pedestrian height, per distance zone on \textit{\gls{vnf} public-test set}.}
    \label{tab:body_height}
    \begin{tabular}{c|c|cc|cc} \hline 
         % Prediction Distance & Total \gls{mjpe} (mm) \\ \hline 

         \multirow{2}{*}{Prediction Distance}& \multicolumn{5}{c}{Body Height Error (cm)}   \\ \cmidrule(lr){2-6}
         & \textbf{Mean Absolute Error} ($\pm$std) & Outdoor & Indoor & Male & Female  \\ \hline 
         
         0 to 5 m & 18.1 ($\pm$ 9.8) & 19.8 & 25.0 & 17.7 & 35 \\ \hline 
         5 to 10 m &  10.5 ($\pm$ 6.5) & 8.7 & 14.8 & 9.4 & 16.2 \\ \hline 
         10 to 15 m & 10.1 ($\pm$ 9.2) & 9.5 & 20.2 & 10.5 & 23.8 \\ \hline 
         15 to 20 m & 21.25 ($\pm$ 11.3) & 21.5 & 31 & 23.3 & 31.5 \\ \hline
    \end{tabular}
\end{table*}

Table \ref{tab:position_error_use_cases} illustrates computed 3D localization errors on on \textit{\gls{vnf} public-test set} and per scenario type, as indoor / outdoor and male / female. When a pedestrian is correctly matched to the ground truth, the \gls{ade} remains consistent across all detection zones and use cases. However, accuracy in the 5–10 m zone is higher than in the 0–5 m zone, likely due to two factors. First, pedestrians in the 5–10 m range are often visible to two cameras, whereas very close pedestrians are typically captured by only one. Second, and likely more impactful, the dataset occasionally contains frames where vehicle doors are open, either blocking the view of a pedestrian entirely or invalidating the vehicle calibration because the left and right cameras are mounted on these doors. Such situations occur exclusively when pedestrians are near the vehicle. We leave it to future users of the dataset to refine measurements in cases where door openings cause these issues.

Table \ref{tab:mjpe} shows the results of skeleton pose estimation in terms of \gls{mjpe}, per various distances to the vehicle. It is visible that the total error increases with the distance, except, again, for the very close zone to the vehicle. The fact that pedestrians appear smaller in the image with distance might explain the increase in error with distance. However, when a pedestrian is very close to the vehicle, some joints are not in the camera \gls{fov}, which might, again, be the reason of the \gls{mjpe} results from 0 to 5 m. Table \ref{tab:mjpe} also illustrates that the 3D-pose estimation is able to perform both at indoor and outdoor and has a similar performances for male and female participants.

Table \ref{tab:body_height} illustrates the error of height estimation. As the actual heights of these persons was measured during the data collection, it was possible to use it as ground truth data. Results shows that the employed 3D-pose estimator's predictions was nearly always taller than the ground truth values (7\% taller in average). It is also visible that the error is more important on female participants, probably linked to the fact that they are shorter in our dataset compared to males. 

%% file: sec/5_conclusion.tex
\section{Conclusion}
\label{sec:conclusion}

In this paper, we introduced the \gls{vnf} dataset, the first publicly available dataset specifically designed to study pedestrian interactions with a stationary vehicle in a near-field perception setting. The dataset includes multi-sensor recordings, featuring \gls{svcs} fisheye camera streams, LiDAR scans, and high-precision 3D skeleton data from a motion capture suit. A subset of this synchronized dataset has been made publicly accessible to facilitate further research in this domain. Additionally, we provided baseline metrics for key pedestrian-related tasks, offering a reference for future studies and benchmarking efforts.

Despite its contributions, the dataset has certain limitations. It primarily focuses on stationary vehicle interactions, leaving dynamic scenarios for future work. Moreover, while the motion capture suit enables precise 3D pose estimation, accumulated drift may introduce minor localization errors over time. Lastly, the dataset's current scope does not extensively cover diverse weather conditions, which is a critical factor in real-world applications. Future iterations of the dataset should aim to include a wider range of environmental conditions to improve the robustness and generalizability of models trained on this data. Despite these limitations, the \gls{vnf} dataset provides a valuable foundation for advancing research in pedestrian perception and safety in autonomous driving systems.

Future research using this dataset could take multiple directions. One avenue is to enhance individual pedestrian perception tasks, such as 2D skeleton detection and tracking, 3D pose uplift, multi-camera 3D localization, and pedestrian trajectory and intent prediction. Alternatively, researchers may explore end-to-end processing pipelines that jointly optimize all these tasks for improved robustness. Furthermore, leveraging synthetic data augmentation or self-supervised learning techniques could help improve model generalization. Ultimately, we hope that this dataset will contribute to advancing pedestrian perception and safety in autonomous driving systems.

%% file: sec/6_Acknowledgments.tex
\section*{Acknowledgment}
We would like to express our gratitude to all anonymous participants who contributed to the data collection process. In addition, we extend our thanks to the dedicated data collection team for their efforts in organizing and recording the collected data.

Part of this work was made possible through the use of the FactoryIA supercomputer, which is financially supported by the Île-de-France Regional Council.